\documentclass[conference]{IEEEtran}
\IEEEoverridecommandlockouts
\usepackage{amsmath,amssymb,amsfonts}
\usepackage[table,xcdraw]{xcolor}
\usepackage{algorithmic}
\usepackage{csquotes}
\usepackage{hyperref}
\usepackage{graphicx}
\usepackage{textcomp}
\usepackage{comment}
\usepackage{url}
\usepackage[style=ieee,citestyle=numeric-comp]{biblatex} 
\usepackage{xcolor}
\usepackage{xskak}
\def\BibTeX{{\rm B\kern-.05em{\sc i\kern-.025em b}\kern-.08em
    T\kern-.1667em\lower.7ex\hbox{E}\kern-.125emX}}

\begin{document}

\title{Supervised and Reinforcement Learning from Observations in Reconnaissance Blind Chess}

\author{\IEEEauthorblockN{Timo Bertram}
\IEEEauthorblockA{\textit{JKU Linz}, Austria} 
\and
\IEEEauthorblockN{Johannes F\"urnkranz}
\IEEEauthorblockA{\textit{JKU Linz}, Austria} 
\and
\IEEEauthorblockN{
Martin M\"uller}
\IEEEauthorblockA{\textit{University of Alberta}, Canada} 
}

\maketitle

\begin{abstract}
In this work, we adapt a training approach inspired by the original AlphaGo system to play the imperfect information game of Reconnaissance Blind Chess. Using only the observations instead of a full description of the game state, we first train a supervised agent on publicly available game records. Next, we increase the performance of the agent through self-play with the on-policy reinforcement learning algorithm Proximal Policy Optimization.
We do not use any search to avoid problems caused by the partial observability of game states and only use the policy network to generate moves when playing. With this approach, we achieve an ELO of 1330 on the RBC leaderboard, which places our agent at position 27 at the time of this writing. We see that self-play significantly improves performance and that the agent plays acceptably well without search and without making assumptions about the true game state.
\end{abstract}

\section{Introduction}

Games have served as immensely popular test domains for artificial intelligence, but the ever-increasing performance in classical board games such as chess and Go has long surpassed human capabilities \cite{silver2016mastering}. However, imperfect information games, where the game state is not perfectly observable, still provide many research challenges for developing competent AI agents. In many of these games, human and AI performance is much closer together \cite{bard2020hanabi,berner2019dota,BrownSuperhuman,brown2020combining,MoravcikStack,vinyals2019grandmaster}, and humans still hold their own in some of these domains. We focus on the game of \emph{Reconnaissance Blind Chess (RBC)}, an imperfect information variant of classical chess, where players only receive limited information about the placement of the opponent's pieces. We aim to apply the training approach used by AlphaGo \cite{silver2016mastering}, which works in perfect information games, to imperfect information games by making some practical adjustments. Specifically, we avoid problems caused by trying to use forward search without having perfect information by only using the trained policy network, which is fully capable of playing the game on its own.

\subsection{Reconnaissance Blind Chess}

In RBC, the game starts with the regular setup of chess pieces. However, a player can only learn about the opponent's moves by a limited form of sensing, which strongly reduces their knowledge of the current state. At the start of each turn, a player senses a $3\times 3$ area of the $8 \times 8$ board, and the true state of these squares is revealed. A player is also informed if their selected move was legal and if they capture a piece. If a move attempts to move through an opponent's piece, the move is truncated to capturing it. Players are also notified whenever one of their own pieces is captured, so they retain perfect information about their own pieces. A game is won by capturing the opponent's king. Finally, check-related chess rules do not apply, so it is legal to castle through a check or even move a king into check. As a consequence, draws are much less common, as stalemate does not exist and even a bare king can still win.

\subsection{Contributions}

Our main contribution is to adapt an AlphaGo-inspired approach \cite{silver2016mastering} to an imperfect information game setting. In AlphaGo, the state is fully observable, so the legal actions of both players are known, which allows deep forward search. In RBC, we generally do not know the true state of the board, which implies that a player cannot know the opponent's options precisely without making significant assumptions beyond the known observations. This greatly restricts the ability to simulate games or conduct a search through a tree of variations. 
In our work, we adapt the early AlphaGo framework, which first primes a neural network by supervised training and then improves it via self-play \cite{silver2016mastering}.
Concretely, we make two main adaptations to account for imperfect information:

\begin{enumerate}
    \item We use the history of observations as our input and avoid any attempt to guess or directly reconstruct the unknown full game state.
    \item We ignore search and solely use the trained policy network to play.
\end{enumerate}

Thus, our network learns to directly map observations to a distribution of actions, which is used to play the game. The aim of this work is to demonstrate that working directly with the given observations of a complex game like RBC, without assumptions about the full hidden game state, is possible and leads to acceptable performance. This opens another angle to work on RBC, which previously strongly focused on trying to explicitly reconstruct the true game state.

\section{Related Work}

Most previous work on RBC is focused on trying to eliminate the uncertainty of RBC, thereby reducing it to normal chess, which then allows the usage of strong search-based chess engines. For example, the runner-up of the 2019 RBC competition, \textit{PiecewiseGrid agent} \cite{highley2020dealing}, maintains a probability distribution over the possible squares for each piece and uses this to compute the likelihood of game states. The program then uses full game states to choose moves with Stockfish \cite{stockfish}, a state-of-the-art chess engine. The 2020 winner \textit{Penumbra} \cite{clark2021deep} does not use a regular chess engine but tries to reduce uncertainty by identifying the opponent, which limits the possible game states, again allowing forward search. Their approach trains a separate network for a list of hand-selected opponents through supervised learning, as well as one catch-all network which is used if the recognition fails. They then generate an approximation of the current state, which is used as the input to the network. However, using opponent-specific training severely limits the flexibility of the approach. The work most similar to ours \cite{savelyev2020mastering} uses an approach similar to AlphaZero. However, their method did not achieve strong performance and barely outperformed a random player. Like many other prior works, they also aimed to reduce uncertainty by trying to identify the most likely game states, which were then used for forward search. 

To the best of our knowledge, there is no previous work on RBC which directly works on the given observations, and we consider this to be the main contribution of our approach. In poker, some previous work exists on directly learning from observations. \cite{DBLP:conf/aaai/YakovenkoCRF16} learned to play simple poker versions from observations of hands, which resulted in a good, but not very strong performance. \cite{DBLP:journals/corr/HeinrichS16} proposed a self-play algorithm that guarantees to converge to a Nash equilibrium. While resulting in similar outcomes, they train a neural network to approximate the average of the past best-responses and we use multiple past agents, leading to a slight difference between their work and the reinforcement learning part of our work. Other strong results in imperfect-information domains are often based on counterfactual regret minimization \cite{DREAM,DeepCFRM,PlayerOfGames}, which may also work well in RBC, but has so far not been explored.

\section{Method}

\begin{table}[b]
\caption{Input representation for the agent (one observation)}
\label{tab:input}
\resizebox{\columnwidth}{!}{%
\begin{tabular}{|l|l|}
\hline
\rowcolor[HTML]{C0C0C0} 
Size of layer & Information represented                          \\ \hline
1             & Square where opponent captured one of our pieces \\ \hline
73            & Last move taken (see \cite{silver2017mastering} for how a move is encoded)                                  \\ \hline
1             & Square where agent captured opponent's piece     \\ \hline
1             & 1 if last move was illegal                       \\ \hline
6             & Position of own pieces (One layer per piece type)                       \\ \hline
1             & Last sense taken                        \\ \hline
6             & Result of last sense (One layer per piece type)                  \\ \hline
1             & Color                                            \\ \hline
\end{tabular}%
}
\end{table}

\begin{figure}
    \centering
    \includegraphics[width = \columnwidth]{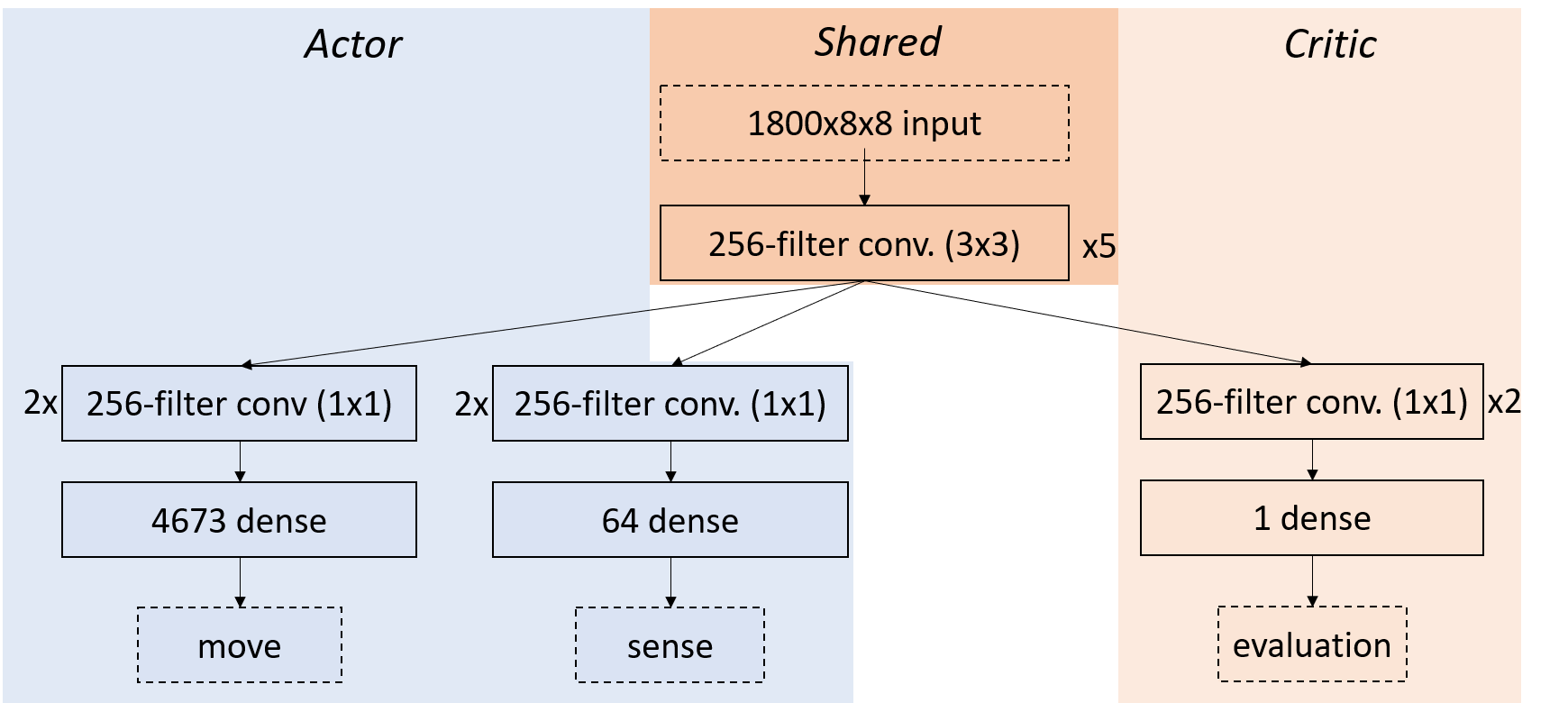}
    \caption{Network architecture}
    \label{fig:architecture}
\end{figure}

In this work, we explicitly aim to not make assumptions about the true game state, but rather learn a policy that directly maps imperfect observations to moves. For this, we represent all information received at each turn to build up a history of observations, which forms the input to our network. The most recent information for a player is represented by a 90 $8\times 8$ bitboard (see Table \ref{tab:input}). Of this, a single 1 in a $73\times 8 \times 8$ stack encodes the last move, which is an idea put forward by \cite{silver2017mastering}. Whether the last move was illegal, and the color of the player, could be represented by a single 0 or 1, but a whole plane is used to facilitate the convolution-based structure of the network. To represent the past, a fixed-size history of the last 20 observations forms the input of size $1800 \times 8 \times 8$ to our network. The network consists of a shared convolutional block, followed by two separate heads for the sense and the move policy (see Figure \ref{fig:architecture}. As a third output, we also tasked the network to predict a scalar outcome of the game, $1$ for win and $-1$ for loss, which was used as the starting network of the critic in reinforcement learning. Although a sequence-based architecture, e.g. a Long Short-Term Memory (LSTM) network or Gated Recurrent Unit (GRU), may intuitively make more sense, we found that in our setup, those networks required more training time without increasing the accuracy of the predictions. Therefore, we speculate that a history of 20 turns is sufficiently long to capture all important information, but even shorter histories could yield benefits by reducing the amount of unnecessary information. We again use $73 \times 8 \times 8$ outputs to describe the proposed move of the network, but we add one more output to represent the option of passing, i.e., to complete the turn without making a move. The sensing policy is modeled as an $8 \times 8$ output, denoting all possible squares on the board. All of our experiments used only a few days of training on a single Tesla M40 and can be reproduced using commonly available hardware.

\subsection{Supervised Learning}

Similar to the setup in \cite{silver2016mastering}, we first train our network on human data before using self-play to tune the policy. To this end,
%
we used a dataset previously used for training the opponent-specific networks of \textit{Penumbra} \cite{clark2021deep}. From these games, we construct training examples by using the taken actions as the target output of the network, with the history of observations as the network's input. In contrast to games like chess and Go, a turn in RBC consists of two actions, sensing and moving. To optimize these two separate but related policies, we use the cross-entropy loss of both heads of the network and optimize their sum. We use all games in the dataset, including losses and games of lower-skilled opponents. Importantly, we do not mask illegal moves, which adds significant difficulty for the network, as only a fraction of all 4673 possible moves is legal in a given position. We also do not mask the 28 outer squares of the board in the sensing policy, which are inferior sensing actions that are dominated by choosing a square in the inner 6x6 area of the board. This decision was made to account for the significant number of sub-optimal senses in the dataset. Future experiments may explore the differences in results when masking those. 

After training for 5 epochs, which is equal to about 8 hours of wall time on our machine, the network achieved a 49.71\% sense and 48.34\% move accuracy on a held-out test set of 10\% of the data, and 50.78\% sensing and 53.91\% moving accuracy on the training data. We stopped training at this point as the network started to overfit. To compute the accuracy, we only counted the outputs of the network which exactly matched the chosen action in the data. A significant number of predictions involve a large degree of randomness, so there is no \enquote{correct} answer for many decisions, but achieving these levels of accuracy in this setting indicates that the data is rather homogeneous. We also counted sensing actions that were strictly better than the sense in the target data as mistakes, such as sensing at g7, which reveals the content of 9 squares, instead of sensing on h8, which only shows a subset of 4 squares. A different measure would have been to count the overlap between the predicted and the target sensing action, but the main problem with this approach is that the purpose of the sensing action is unknown. For example, its true intent might have been collecting information about one particular square only.

\subsection{Reinforcement Learning}

Although with our agent achieved a good accuracy of predictions through supervised training, it did not learn learn to actually win the game. Supervised learning attributes the same importance to correctly predicting to capture the opponent's king (which wins the game) as to playing the opening move 1.\hspace{1mm}e4. Moreover, it may also learn low-quality in-game actions such as sensing at the edge of the board, which is never an optimal decision. Since the goal of the agent should be to win the game, the second stage of training by reinforcement learning rewards the agent only for achieving this objective. 
We frame the problem as a Markov decision problem (MDP), where the opponent is part of the stochastic environment, and use an on-policy reinforcement learning algorithm. Although RBC is a partially observable Markov decision problem (POMDP), we disregard the partial observability of the domain and use the history of observations as if they were a complete description of the state, thus approximating the POMDP as an MDP. 
\medskip

We train the agent by self-play and collect the actions taken and the result of each game. With this experience, we optimize our agent using our own implementation of Proximal Policy Optimization (PPO) \cite{schulman2017proximal}. To simultaneously optimize both policies (sensing and moving) we compute separate losses for both of them and perform gradient descent on their sum. While sensing is a passive action that never directly results in winning the game, we reward the final winning move and the sense just before it with a reward of 1. 
To reduce overfitting to the current version of the agent, we let it play against randomly selected past versions of itself, which are saved whenever they reach a 65\% win rate. This is similar to playing against the average strategy of past best-responses, and a well-known technique \cite{vinyals2019grandmaster,DREAM,DeepCFRM}.

In training, the probability $p_i$ of playing against a version $i$ of the network depends on the win-rate $w_i$ of the training against it:

\begin{equation}
    p_i = 1-(w_i/ \sum_{i}w_i)
\end{equation}

The win-rate $w_i$ is approximated as the average of the $K=500$ most recent results of the training agent against it. 

One important consideration in RBC, and an important difference from normal chess, is that due to the partial observability, the optimal strategy should be stochastic, as a deterministic strategy can easily be exploited. However, when testing our agent, we found that choosing the action with highest probability lead to slightly better performance.

\section{Results}

We tested the performance of our agent at two time points; after supervised training and after reinforcement learning. The hypothesis is that adding reinforcement learning, which directly aims to optimize the real objective of winning games, should increase the win-rate compared to only supervised training on public games. 

\subsection{ELO Performance}

In order to evaluate the performance, we uploaded both versions to the public leaderboard\footnote{\url{https://rbc.jhuapl.edu/}}, which results in an approximate ELO rating for each agent. 
After the supervised training, our agent's performance is similar to that of the publicly available baseline agent \textit{Trout}, which is a naive Stockfish-based agent. However, as seen in Table~\ref{tab:performance}, reinforcement learning leads to a tangible performance benefit on the leaderboard. In training, observed that it consistently learned to win against its previous version. For example, the reinforced agent exhibits a win rate of more than 80\% against the supervised agent throughout training. At the time of this writing, these results put our agent at rank 27 out of 84 on the leaderboard, although we did not see an indication of convergence.

\begin{table}[h]
\caption{Performance comparison of the proposed agent}
\label{tab:performance}
\centering
\begin{tabular}{|l|c|}
\hline
\rowcolor[HTML]{C0C0C0} 
{\color[HTML]{333333} Name} & {\color[HTML]{333333} Performance} \\ \hline
Supervised agent            & 1118                               \\ \hline
Reinforced agent            & 1330                               \\ \hline
Trout (Public Stockfish baseline)  & 1111                               \\ \hline
\end{tabular}%
\end{table}

\subsection{Analysis of Example Games}

We observed that our agent has a highly aggressive, even reckless at times, playing style. It often aims for very quick attacks on the enemy king, sacrificing one or more pieces (including very early queen sacrifices) in order to get to a position where the opponent's king is no longer surrounded by defenders and is not able to defend reliably against multiple possible and unobserved threats of the agent. This kind of strategy works well against many lower and middle-skilled opponents, and even scores the occasional win against top contenders.

One game we want to highlight can be replayed at \url{https://rbc.jhuapl.edu/games/462287}, where the agent played against one of the higher-rated opponents on the leaderboard. In the game, our agent created two situations where the opponent could not certainly determine from which square its king was attacked but was able to sense the correct positions in order to defend against the threats. In contrast, playing against a lower-rated opponent (\url{https://rbc.jhuapl.edu/games/462288}) the same strategy worked well, as the opponent did not have information about the bishop on c4, which lead to a quick win. Similarly, in \url{https://rbc.jhuapl.edu/games/462249}, our agent continuously made threats, which in the end led to an undetected knight capturing the king. Such a strategy would not work well in classical chess, which provides some evidence that policies in chess and RBC are not necessarily similar, and that trying to reduce RBC to chess may be problematic at times.


\section{Conclusion}

In this work, we show our first results on applying an AlphaGo-inspired training regime to the imperfect information game of Reconnaissance Blind Chess. Our agent learns to use a history of observations to create distributions of actions for both sensing and playing. First, we use supervised training on publicly available expert games, where the task is to predict the actions of the experts. Next, we use on-policy reinforcement learning with self-play to strengthen the playing performance of the agent. With this approach, we reached rank 27 of 84 on the leaderboard and an estimated ELO of 1330, using no further game-specific optimizations.

To continue this work, we aim to refine our self-playing process. It is currently unclear whether this process alone can lead to top performance.. Incorporating experience gained from playing on the leaderboard is much slower than playing games against itself offline but may lead to more valuable information from varied strong opponents, thus facilitating quicker improvement. An additional angle that we aim to tackle is a combination of the trained agent with a classical engine like Stockfish. Combining action suggestions from both, or adapting Stockfish's moves by using the probability distribution of the agent, can lead to a more normal and classical playing style, while also using learned experience from RBC self-play.

\section*{Acknowledgements}
We thank the reviewers of this paper for providing excellent feedback on improving the presentation, pointers to related works that we had missed, and suggestions for continuing this line of work.

\printbibliography

@article{silver2016mastering,
  title={Mastering the game of {G}o with deep neural networks and tree search},
  author={Silver, David and Huang, Aja and Maddison, Chris J and Guez, Arthur and Sifre, Laurent and Van Den Driessche, George and Schrittwieser, Julian and Antonoglou, Ioannis and Panneershelvam, Veda and Lanctot, Marc and others},
  journal={Nature},
  volume={529},
  number={7587},
  pages={484--489},
  year={2016},
  publisher={Nature Publishing Group}
}

@article{DREAM,
  author    = {Eric Steinberger and
               Adam Lerer and
               Noam Brown},
  title     = {{DREAM:} {D}eep Regret minimization with Advantage baselines and Model-free
               learning},
  journal   = {CoRR},
  volume    = {abs/2006.10410},
  year      = {2020},
}

@article{DeepCFRM,
  author    = {Eric Steinberger},
  title     = {Single Deep Counterfactual Regret Minimization},
  journal   = {CoRR},
  volume    = {abs/1901.07621},
  year      = {2019}
}

@article{PlayerOfGames,
  author    = {Martin Schmid and
               Matej Moravcik and
               Neil Burch and
               Rudolf Kadlec and
               Joshua Davidson and
               Kevin Waugh and
               Nolan Bard and
               Finbarr Timbers and
               Marc Lanctot and
               Zach Holland and
               Elnaz Davoodi and
               Alden Christianson and
               Michael Bowling},
  title     = {Player of Games},
  journal   = {CoRR},
  volume    = {abs/2112.03178},
  year      = {2021}
}

@article{vinyals2019grandmaster,
  title={Grandmaster level in {StarCraft II} using multi-agent reinforcement learning},
  author={Vinyals, Oriol and Babuschkin, Igor and Czarnecki, Wojciech M and Mathieu, Micha{\"e}l and Dudzik, Andrew and Chung, Junyoung and Choi, David H and Powell, Richard and Ewalds, Timo and Georgiev, Petko and others},
  journal={Nature},
  volume={575},
  number={7782},
  pages={350--354},
  year={2019},
  publisher={Nature Publishing Group}
}

@inproceedings{DBLP:conf/aaai/YakovenkoCRF16,
  author    = {Nikolai Yakovenko and
               Liangliang Cao and
               Colin Raffel and
               James Fan},
  title     = {Poker-CNN: {A} Pattern Learning Strategy for Making Draws and Bets
               in Poker Games Using Convolutional Networks},
  booktitle = {Proceedings of the 30th {AAAI} Conference on Artificial Intelligence},
  pages     = {360--368},
  publisher = {{AAAI} Press},
  year      = {2016}
}

@article{brown2020combining,
  title={Combining deep reinforcement learning and search for imperfect-information games},
  author={Brown, Noam and Bakhtin, Anton and Lerer, Adam and Gong, Qucheng},
  journal={Advances in Neural Information Processing Systems},
  volume={33},
  pages={17057--17069},
  year={2020}
}

@article{berner2019dota,
  author    = {Christopher Berner and
               Greg Brockman and
               Brooke Chan and
               Vicki Cheung and
               Przemyslaw Debiak and
               Christy Dennison and
               David Farhi and
               Quirin Fischer and
               Shariq Hashme and
               Christopher Hesse and
               Rafal J{\'{o}}zefowicz and
               Scott Gray and
               Catherine Olsson and
               Jakub Pachocki and
               Michael Petrov and
               Henrique Pond{\'{e}} de Oliveira Pinto and
               Jonathan Raiman and
               Tim Salimans and
               Jeremy Schlatter and
               Jonas Schneider and
               Szymon Sidor and
               Ilya Sutskever and
               Jie Tang and
               Filip Wolski and
               Susan Zhang},
  title     = {Dota 2 with Large Scale Deep Reinforcement Learning},
  journal   = {CoRR},
  year      = {2019},
  eprinttype = {arXiv},
  eprint    = {1912.06680},
  bibsource = {dblp computer science bibliography, https://dblp.org}
}

@article{highley2020dealing,
  title={Dealing with uncertainty: A piecewise grid agent for reconnaissance blind chess},
  author={Highley, Timothy and Funk, Brendan and Okin, Laureen},
  journal={Journal of Computing Sciences in Colleges},
  volume={35},
  number={8},
  pages={156--165},
  year={2020},
  publisher={Consortium for Computing Sciences in Colleges}
}

@misc{stockfish,
  author = {Romstad, Tord and Costalba, Marco and Kiiski, Joona},
  title = {Stockfish},
  howpublished = "\url{https://stockfishchess.org/}",
  year = {2022}, 
  note = "[accessed 26-April-2022]"
}

@article{clark2021deep,
  title={Deep Synoptic {Monte-Carlo} Planning in Reconnaissance Blind Chess},
  author={Clark, Gregory},
  journal={Advances in Neural Information Processing Systems},
  volume={34},
  year={2021}
}

@article{silver2017mastering,
  author    = {David Silver and
               Thomas Hubert and
               Julian Schrittwieser and
               Ioannis Antonoglou and
               Matthew Lai and
               Arthur Guez and
               Marc Lanctot and
               Laurent Sifre and
               Dharshan Kumaran and
               Thore Graepel and
               Timothy P. Lillicrap and
               Karen Simonyan and
               Demis Hassabis},
  title     = {Mastering Chess and Shogi by Self-Play with a General Reinforcement
               Learning Algorithm},
  journal   = {CoRR},
  year      = {2017},
  eprinttype = {arXiv},
  eprint    = {1712.01815}
}

@article{schulman2017proximal,
  author    = {John Schulman and
               Filip Wolski and
               Prafulla Dhariwal and
               Alec Radford and
               Oleg Klimov},
  title     = {Proximal Policy Optimization Algorithms},
  journal   = {CoRR},
  year      = {2017},
  eprinttype = {arXiv},
  eprint    = {1707.06347}
}

@thesis{savelyev2020mastering,
  author       = {Sergey Savelyev}, 
  title        = {Mastering Reconnaissance Blind Chess with Reinforcement Learning},
  school       = {Georgia Institute of Technology},
  year         = 2020,
  type = {Undergraduate Thesis}
}

@article{bard2020hanabi,
  title={The {H}anabi challenge: A new frontier for {AI} research},
  author={Bard, Nolan and Foerster, Jakob N and Chandar, Sarath and Burch, Neil and Lanctot, Marc and Song, H Francis and Parisotto, Emilio and Dumoulin, Vincent and Moitra, Subhodeep and Hughes, Edward and others},
  journal={Artificial Intelligence},
  volume={280},
  year={2020},
  publisher={Elsevier}
}

@article{MoravcikStack,
author = {Matej Moravčík  and Martin Schmid  and Neil Burch  and Viliam Lisý  and Dustin Morrill  and Nolan Bard  and Trevor Davis  and Kevin Waugh  and Michael Johanson  and Michael Bowling },
title = {DeepStack: Expert-level artificial intelligence in heads-up no-limit poker},
journal = {Science},
volume = {356},
number = {6337},
pages = {508-513},
year = {2017}}

@article{BrownSuperhuman,
author = {Noam Brown  and Tuomas Sandholm },
title = {Superhuman {AI} for multiplayer poker},
journal = {Science},
volume = {365},
number = {6456},
pages = {885-890},
year = {2019}}

@article{DBLP:journals/corr/HeinrichS16,
  author    = {Johannes Heinrich and
               David Silver},
  title     = {Deep Reinforcement Learning from Self-Play in Imperfect-Information
               Games},
  journal   = {CoRR},
  year      = {2016},
  eprinttype = {arXiv},
  eprint    = {1603.01121}
}

\end{document}